%% file: root.tex
\documentclass[letterpaper,10pt,journal,twoside]{IEEEtran}

\usepackage{amsmath,amsfonts,amssymb}
\usepackage{array}
\usepackage{textcomp}
\usepackage{threeparttable}
\usepackage{verbatim}
\usepackage{graphicx}
\usepackage{cite}
\usepackage[utf8]{inputenc}
\usepackage[T1]{fontenc}
\usepackage{hyperref}
\usepackage{url}
\usepackage{booktabs}
\usepackage{nicefrac}
\usepackage{microtype}
\usepackage{multirow}
\usepackage{siunitx}
\usepackage{tabularx}
\usepackage{float}
\usepackage{stfloats}
\usepackage{xcolor}

\hypersetup{hypertexnames=false}
\title{Unleashing the Agility of Wheeled-Legged Robots for High-Dynamic Reflexive Obstacle Evasion}

\author{Yongen Zhao$^{1,5}$, Zihao Xu$^{2}$, Wenzhi Lu$^{1}$, Zhen Chu$^{3}$,
        Kailin Lyu$^{4,5}$, Hao Sun$^{6}$, and Ce Hao$^{5,*}$%
\thanks{Manuscript received April 15, 2026; accepted September 19, 2026. This paper has been accepted for publication in \textit{IEEE Robotics and Automation Letters (RA-L)}. \textit{(Corresponding author: Ce Hao$^{*}$.)}}%
\thanks{This work was supported by the Zhongguancun Academy (Grant No. c20250502). We would like to thank DEEP Robotics (Hangzhou, China) for hardware
        support and technical assistance with the M20 platform.}%
\thanks{$^{1}$Yongen Zhao and Wenzhi Lu are with the School of Mechanical Engineering, Tianjin University, Tianjin 300072, China.}%
\thanks{$^{2}$Zihao Xu is with the School of Computing, National University of Singapore, Singapore 117417.}%
\thanks{$^{3}$Zhen Chu is with Deep Robotics, Hangzhou 311000, China.}%
\thanks{$^{4}$Kailin Lyu is with the Institute of Automation, Chinese Academy of Sciences, Beijing 100190, China.}%
\thanks{$^{5}$Yongen Zhao, Kailin Lyu, and Ce Hao are with Zhongguancun Academy, Beijing 100094, China (e-mail: \href{mailto:haoce@bza.edu.cn}{haoce@bza.edu.cn}).}%
\thanks{$^{6}$Hao Sun is with Guangzhou Automobile Group Co., Ltd., Guangzhou 511434, China.}%
}

\begin{document}

\maketitle

\input{sections/0_Abstract}


\IEEEpeerreviewmaketitle

\input{sections/1_Introduction}
\input{sections/2_Related_works}
\input{sections/3_Preliminary}
\input{sections/4_Method}
\input{sections/5_Experiment}
\input{sections/6_Conclusion}

\bibliographystyle{IEEEtran}
\bibliography{references}

\end{document}

%% file: sections/0_Abstract.tex
\begin{abstract}
Wheeled-legged robots combine the efficiency of rolling with the adaptability of legged locomotion, offering unique agility for dynamic environments. However, enabling rapid reflexive evasion remains challenging due to the coexistence of heterogeneous wheel–leg dynamics, hybrid locomotion modes, and non-holonomic constraints. In this work, we investigate how wheeled-legged robots can exploit their hybrid morphology for high-dynamic obstacle avoidance. We propose AWARE, a hierarchical reinforcement learning framework that decomposes avoidance into navigation-avoidance and reflexive-evasion regimes coordinated by a threat-conditioned high-level policy. By learning specialized low-level experts, AWARE autonomously discovers distinct rolling-, stepping-, and hybrid-dominated evasive behaviors, including forward lunges and lateral dodges. Simulation experiments across different reaction times and approach directions, together with real-robot evaluations on the M20 platform, demonstrate improved evasion capability and effective online transition between locomotion regimes. These results highlight the potential of exploiting hybrid wheel–leg actuation for agile and reactive mobility in dynamic environments. Paper homepage: \url{https://aware-ral-2026.github.io/}.
\end{abstract}

%% file: sections/1_Introduction.tex
\section{Introduction} \label{Sec: intro}

\begin{figure*}[tbp]
    \centering
    \includegraphics[width=0.65\textwidth]{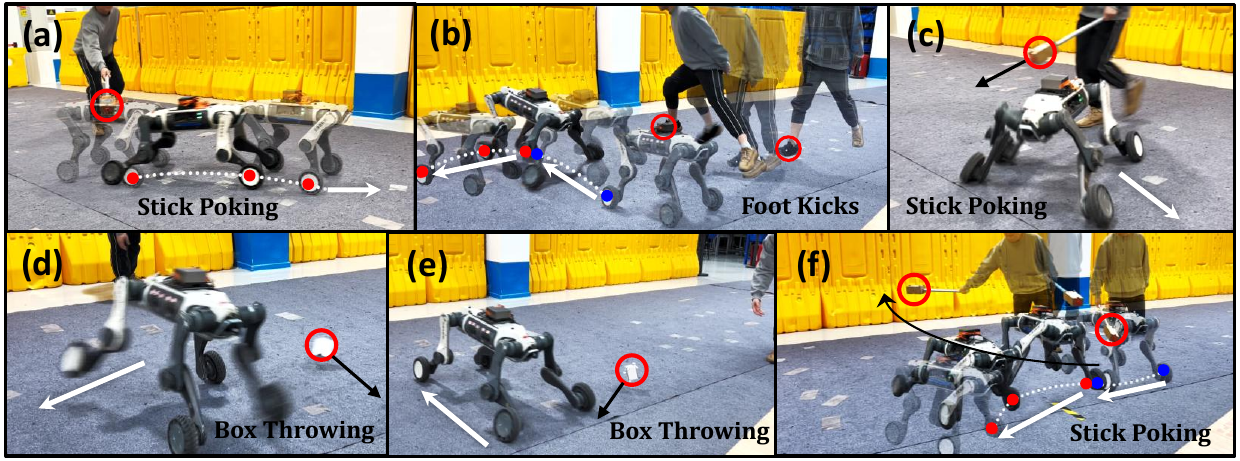}
    \caption{\textbf{AWARE: Adaptive Wheeled-Legged Avoidance and Reflexive Evasion.} AWARE enables wheeled-legged robots to execute reflexive evasion and mixed-mode avoidance behaviors, with online transitions between navigation-avoidance and reflexive-evasion. The figure shows representative real-robot interactions involving stick poking, foot kicks, and box throwing from different approach directions.}
    \label{Fig: teaser}
    \vspace{-5mm}
\end{figure*}

The integration of wheeled efficiency with legged versatility has established wheeled-legged robots \cite{zhou2023max, schwarz2016hybrid, wang2021balance, klemm2019ascento, kashiri2019centauro} as highly capable platforms for navigating complex, unstructured environments. By combining rapid rolling with terrain-adaptive stepping, these platforms provide a broader locomotion repertoire than purely wheeled~\cite{shih2023ackerman} or purely legged systems \cite{lee2024learning}. Consequently, they are increasingly deployed in real-world applications ranging from industrial inspection to urban delivery\cite{zhou2023max, sun2025atros}. 
Despite these significant advancements, deploying wheeled-legged robots in increasingly dense and unpredictable human-centric environments motivates rapid reflexive responses to fast-moving dynamic obstacles.

However, achieving high-dynamic obstacle avoidance for wheeled-legged robots remains challenging because rolling and stepping impose different dynamical properties and control constraints. This hybrid embodiment raises a specific control challenge: how to coordinate navigation-avoidance and reflexive-evasion under hybrid dynamics and non-holonomic constraints. Traditional optimization-based methods, such as Model Predictive Control (MPC)~\cite{vicente2022fast}, are well-suited for wheeled-legged robots \cite{bjelonic2020rolling,hosseini2023dynamic,bjelonic2019keep,li2022balancing} and quadrupedal robots \cite{chiu2022collision,grandia2023perceptive,gaertner2021collision} but often incur prohibitive computational costs when handling the high-dimensional, whole-body kinematics required for agile evasion. Existing research on agile navigation and locomotion has predominantly focused on purely quadrupedal\cite{xu2025rebot, he2024agile, zhang2024learning,miki2022learning,hwangbo2019learning, rudin2022advanced}, UAVs\cite{zhou2020ego, xu2025intent}, mobile manipulators\cite{burgess2024reactive,chiu2022collision} domains.
In particular, REBot~\cite{xu2025rebot} formulates a reflexive evasion problem for legged systems, but does not account for the unique hybrid dynamics and high-speed rolling behavior of wheeled-legged robots. This leaves the conflict between navigation-avoidance and reflexive-evasion insufficiently addressed.

To investigate the agility potential and behavioral characteristics of wheeled-legged robots in reflexive evasion tasks, we propose the \textbf{Adaptive Wheeled-Legged Avoidance and Reflexive Evasion (AWARE)} framework (Fig.~\ref{Fig: teaser}), a hierarchical reinforcement learning architecture for high-dynamic obstacle avoidance. The key idea is to decouple the problem into a high-level evasive policy and a low-level dual-mode locomotion controller, where the high-level policy dynamically routes velocity commands to one of two specialized pre-trained experts: a low-speed omnidirectional network for navigation avoidance and a high-dynamic agile network for reflexive evasion. We further analyze the emergent gait modes and evasion modes under different obstacle conditions.

Simulation evaluations in Isaac Lab show that AWARE produces distinct evasive behaviors, including forward lunges and lateral dodges. In real-world deployment on the M20 wheeled-legged robot, box throwing, stick poking, and foot kicks are used to emulate highly dynamic obstacle interactions. Real-robot experiments on the M20 platform further validate the feasibility of executing and switching between the learned evasive behaviors under representative dynamic interactions.

The core contributions of this paper are summarized as follows:
\begin{itemize}
    \item We formulate reflexive evasion for wheeled-legged robots as a constrained control problem under short reaction windows, hybrid wheel--leg dynamics, and non-holonomic constraints.
    \item We propose AWARE, a dual-regime framework with specialized low- and high-dynamic experts and learned threat-conditioned routing.
    \item We validate AWARE in simulation and on a real-robot platform, demonstrating short-reaction-time evasion, distinct wheel--leg behaviors, and adaptive mode transitions.
\end{itemize}

%% file: sections/2_Related_works.tex
\section{Related Work} \label{Sec: related}

\subsection{Dynamic and Reflexive Obstacle Evasion}

Dynamic obstacle evasion is commonly addressed through optimization-based planning~\cite{zhang2025dynamic} or learning-based reactive control. Model predictive control methods explicitly predict obstacle motion and optimize collision-free whole-body trajectories under dynamic constraints \cite{vicente2022fast,chiu2022collision,xu2025intent,grandia2023perceptive,gaertner2021collision}, but their effectiveness under short reaction windows depends on model accuracy and online optimization efficiency. Learning-based methods instead provide high-frequency reactive control and have demonstrated agile locomotion and obstacle avoidance on legged robots \cite{xu2025rebot,he2024agile,zhang2024learning,miki2022learning,hwangbo2019learning,hoeller2024anymal}. In particular, REBot \cite{xu2025rebot} learns reflexive avoidance and recovery behaviors for rapidly approaching obstacles. However, these methods are primarily developed around a single leg-dominated execution regime and do not explicitly address the coexistence of rolling-, stepping-, and braking-dominated dynamics in wheeled-legged systems. This creates interference between navigation-avoidance and reflexive-evasion when represented by a unified controller.

\subsection{Wheeled-Legged Locomotion and Skill-Specialized Control}

Existing wheeled-legged methods use trajectory optimization, whole-body control, and reinforcement learning to achieve balance, hybrid locomotion, terrain traversal, and autonomous navigation \cite{zhou2023max,schwarz2016hybrid,wang2021balance,klemm2019ascento,kashiri2019centauro,bjelonic2020rolling,hosseini2023dynamic,bjelonic2019keep,li2022balancing,sun2025atros,lee2024learning}. Related learning-based approaches further improve multi-skill control through hierarchical policies, motion specialization, or mixture-of-experts architectures \cite{rudin2022advanced,chen2025gmt,huang2025moe}. These methods mainly target navigation-avoidance quality, terrain-dependent skill composition, or task-level navigation, rather than reflexive responses under severely limited reaction time. In contrast, AWARE introduces a dual-regime decomposition that separates navigation-avoidance and reflexive-evasion into specialized experts, while a threat-conditioned high-level policy jointly generates evasive commands and selects the corresponding execution regime.

%% file: sections/3_Preliminary.tex
\section{Preliminaries} \label{Sec: prelim}

To simplify collision detection and enhance computational efficiency, we model the currently active obstacle $\mathcal{O}_o$ using a spherical approximation. The core of safe avoidance lies in finding a control law $\pi$, subject to dynamic constraints, that ensures the robot's collision bounding box $\mathcal{B}(q)$ and the active obstacle $\mathcal{O}_o$ satisfy the safety invariant condition:
$
\mathrm{dist}\big( \mathcal{B}(q),  \mathcal{O}_o \big) \geq d_{\text{safe}},
\label{eq:safety}
$
where $d_{\text{safe}}$ is a predefined safety margin.

The obstacle avoidance task is formally defined as an optimal control problem under time-varying constraints for a single active dynamic threat. At each time step, the system receives observations from both the robot and the environment, as summarized in Table \ref{tab:system_inputs}. The goal is to find an optimal control policy $\pi^*$ that generates whole-body control commands $\mathbf{u}(t)$ (wheel velocities and joint positions) such that the robot's physical bounding box $\mathcal{B}(q)$ remains disjoint from the spatial region occupied by the active obstacle $\mathcal{B}_o(t)$ throughout the finite evasion horizon $[0,T]$.

\begin{table}[t]
\centering
\caption{SYSTEM VARIABLES FOR OBSTACLE AVOIDANCE}
\label{tab:system_inputs}
\begin{tabular}{lll}
\toprule
\textbf{Category} & \textbf{Notation} & \textbf{Description} \\
\midrule
\multirow{5}{*}{\textbf{Robot State}} 
& $\mathbf{p}_r \in \mathbb{R}^3$ & Base position  \\
& $\mathbf{R}_r \in \mathrm{SO}(3)$ & Base orientation  \\
& $\boldsymbol{\omega}_r \in \mathbb{R}^3$ & Base angular velocity \\
& $\mathbf{q}, \dot{\mathbf{q}} \in \mathbb{R}^{n_j}$ & Joint positions and velocities \\
\midrule
\multirow{4}{*}{\textbf{Obstacle Information}} 
& $\mathbf{p}_o \in \mathbb{R}^3$ & Position of the active obstacle \\
& $\mathbf{v}_o \in \mathbb{R}^3$ & Velocity of the active obstacle \\
& $r_o \in \mathbb{R}$ & Radius of the active obstacle  \\
& $d_{\text{safe}} \in \mathbb{R}$ & Predefined safety margin \\
\midrule
\textbf{Control Output} & $\mathbf{u}(t)$ & joint and wheel commands \\
\bottomrule

\end{tabular}
\rule{0pt}{0mm}
\end{table}

To quantitatively evaluate the effectiveness of a single avoidance trial, a trial is deemed a Success if and only if the robot simultaneously satisfies the following core conditions:
\begin{itemize}
\item \textbf{Collision-free Safety}: No part of the robot body comes into contact with the active obstacle throughout the entire time horizon $[0, T]$.
\item \textbf{Dynamic Stability}: During high-maneuver actions (such as dashing or jumping), the robot does not experience rollover or loss of control.
\end{itemize}

The framework proposed in this paper aims to intelligently decide and generate corresponding control commands $\mathbf{u}(t)$ based on the real-time sensed active-obstacle state $s_o = [\mathbf{p}_o, \mathbf{v}_o, r_o]^\top$, achieving a precise avoidance response.

%% file: sections/4_Method.tex
 \section{Method} \label{Sec: method}

\begin{figure*}[t]
    \centering
    \includegraphics[width=0.70\textwidth]
    {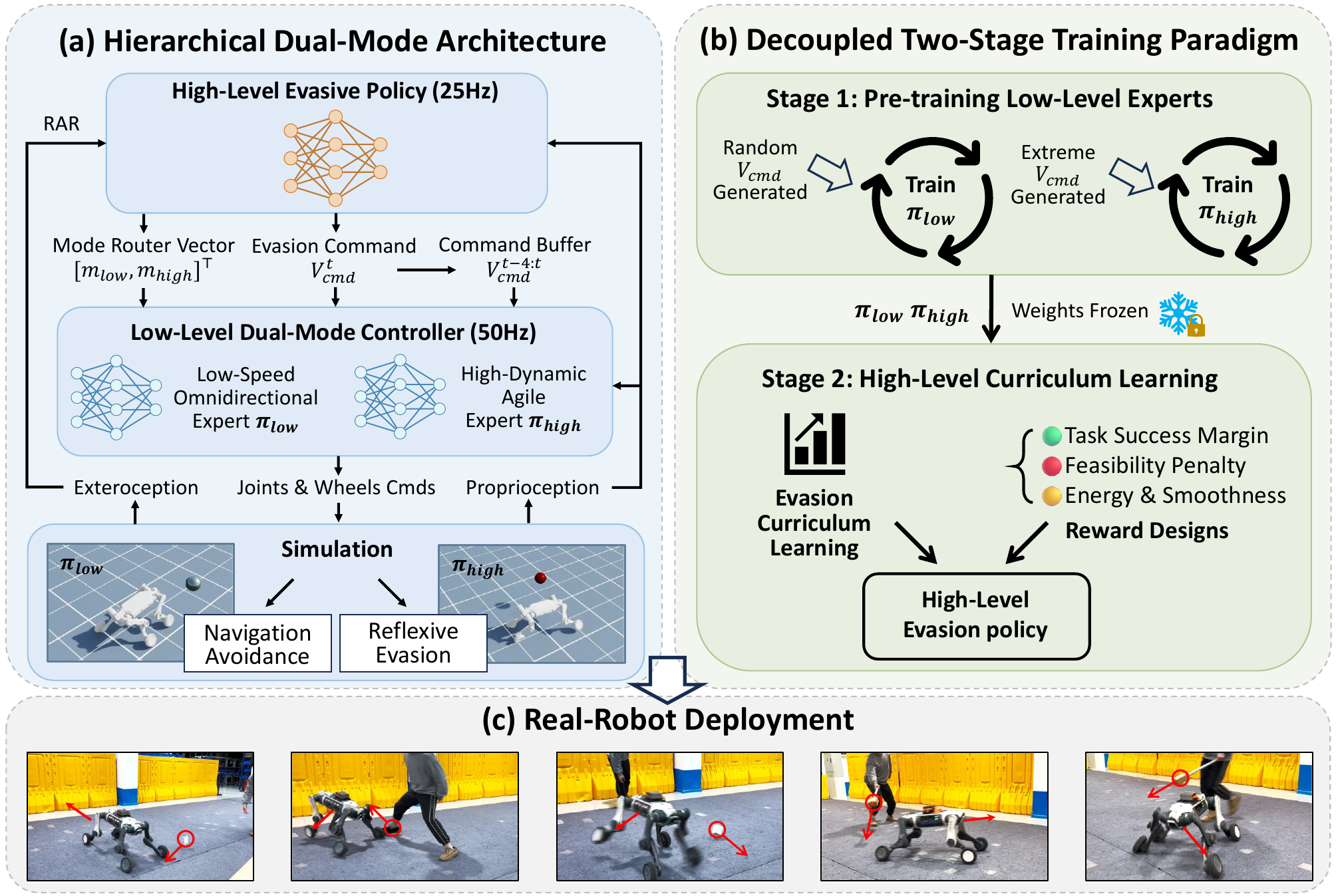}
    \caption{Overview of the AWARE framework. \textbf{(a) Hierarchical Architecture:} A high-level policy generates evasion commands that are executed by specialized low-level experts for either smooth navigation avoidance or high-dynamic reflexive evasion. \textbf{(b) Two-Stage Training:} A decoupled pipeline sequentially trains the low-level experts and the high-level policy to optimize overall evasion performance and efficiency. \textbf{(c) Real-Robot Deployment:} We deploy our system on the M20 robot platform under representative dynamic-obstacle interactions.}
    \label{Fig: pipeline}
\vspace{-5mm}
\end{figure*}
In this section, we introduce our \textbf{Adaptive Wheeled-legged Avoidance and Reflexive Evasion (AWARE)} framework in subsection \ref{subsec: Hierarchical Adaptive Avoidance Architecture} and the two-stage training paradigm adopted in subsection \ref{subsec: Two-Stage Training and Curriculum Learning} (Fig. \ref{Fig: pipeline}).

\subsection{Hierarchical Adaptive Avoidance Architecture}
\label{subsec: Hierarchical Adaptive Avoidance Architecture}
The \textbf{AWARE} framework decouples the problem into two distinct modules: a High-Level Evasive Policy that implicitly plans avoidance maneuvers, and a Low-Level Dual-Mode Locomotion Controller that executes specialized evasion skills.
The AWARE framework is activated based on the relative obstacle state\cite{xu2025rebot}. Specifically, when the relative obstacle velocity and relative distance satisfy a predefined threat condition,
\begin{equation}
\alpha_t = \mathbb{I}\!\left(\|\mathbf{v}_{\mathrm{rel},t}\| > v_{\mathrm{th}} \;\land\; \|\mathbf{p}_{\mathrm{rel},t}\| < d_{\mathrm{th}}\right),
\label{eq:threat_activation}
\end{equation}

\noindent the system triggers the evasive policy and generates an appropriate sequence of avoidance actions.
We set $v_{\mathrm{th}}=0.5\,\mathrm{m/s}$ and $d_{\mathrm{th}}=3\,\mathrm{m}$ in all experiments. These thresholds define a conservative supervisory activation condition during deployment.

\begin{figure}[!t]
    \centering
    \includegraphics[width=0.85\columnwidth]{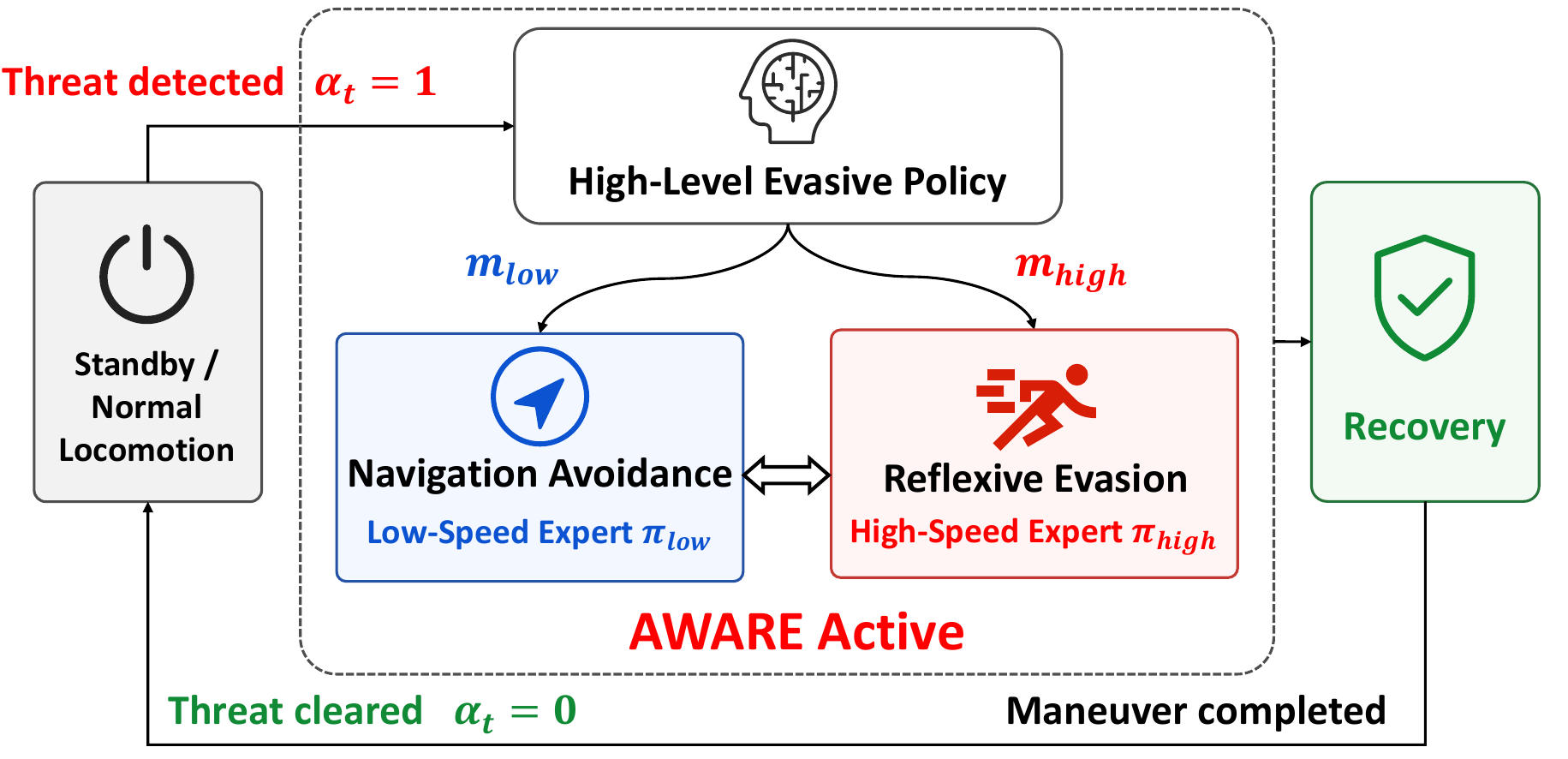}
    \vspace{-2mm}
    \caption{Mechanism of AWARE reflexive evasion. Relative obstacle states activate the reflexive module; the high-level policy estimates threat severity, outputs an evasive velocity command, and routes execution to the navigation-avoidance or reflexive-evasion expert.}
    \label{Fig: mechanism}
    \vspace{-2mm}
\end{figure}

Fig.~\ref{Fig: mechanism} further details the online execution mechanism of AWARE. The activation logic filters obstacle states according to relative speed and distance, while the high-level policy uses the RAR feature and proprioceptive state to jointly determine the evasive command and the execution mode. This separates threat assessment from physical execution, allowing the low-level experts to focus on their respective locomotion regimes.

\textbf{High-Level Evasive Policy.}
The upper-level network functions as an intelligent maneuver planner and mode selector. To explicitly quantify the threat level of dynamic obstacles, we introduce the \textit{Relative Approaching Rate} (RAR), denoted as $\kappa$
, as a crucial guiding feature, 
defined as the negative time derivative of the relative distance:
\begin{equation}
\small
\kappa = -\frac{d}{dt} |\mathbf{p}_{\text{rel}}| = -\frac{\mathbf{p}_{\text{rel}}^\top \mathbf{v}_{\text{rel}}}{|\mathbf{p}_{\text{rel}}|}.
\label{eq:kappa}
\end{equation}
Under this convention, $\kappa>0$ indicates that the obstacle is closing, whereas $\kappa<0$ indicates that it is moving away in the radial direction.

The upper-level policy takes the computed RAR value $\kappa$, the real-time relative position of the obstacle $\mathbf{p}_{\text{rel}}$, and the robot's current proprioceptive state $\mathbf{x}^t$ as its observation. 
The policy implicitly plans evasive maneuvers by outputting a desired velocity command $v_{\mathrm{cmd}} \in \mathbb{R}^3$ at each time step.
It also explicitly selects an execution mode, dictating the activation of the subsequent low-level expert networks. From the last hidden layer of the upper-level neural network, we branch out a discrete mode selector implemented via a Gumbel-Softmax distribution to generate a one-hot vector $\mathbf{m} = [m_{\text{low}}, m_{\text{high}}]^\top \in \{0, 1\}^2$.

\textbf{Low-Level Dual-Mode Locomotion Controller.} The lower-level module acts as the physical executor, translating the commands into joint positions and wheel velocities. The observation space of this lower-level policy is constructed by concatenating the current velocity command $\mathbf{v}^t_{\text{cmd}}$, the historical commands from the previous four time steps $\mathbf{v}^{t-4:t-1}_{\text{cmd}}$, and the robot's real-time proprioceptive state $\mathbf{x}^t$. 

Unlike existing frameworks that often employ teacher-student architectures ~\cite{miki2022learning,lee2024learning} or Mixture-of-Experts (MoE) ~\cite{huang2025moe,chen2025gmt} to expand policy expressiveness, we construct the low-level controller using two specialized expert networks ~\cite{hoeller2024anymal} to support reflexive evasion under high-threat scenarios.
The final whole-body action command $\mathbf{a}^t$ is deterministically routed by the one-hot vector $\mathbf{m}$ output from the high-level policy:
\begin{equation}
\small
\mathbf{a}^t = m_{\text{low}} \cdot \pi_{\text{low}}(\mathbf{v}_{\text{cmd}}^{t-4:t}, \mathbf{x}^t) + m_{\text{high}} \cdot \pi_{\text{high}}(\mathbf{v}_{\text{cmd}}^{t-4:t}, \mathbf{x}^t).
\label{eq:action_routing}
\end{equation}
Low-Speed Omnidirectional Network ($\pi_{\text{low}}$) is dedicated to precise tracking of $\mathbf{v}_{\text{cmd}}$ in less urgent scenarios. 
The High-Dynamic Agile Network ($\pi_{\text{high}}$) is selected through the learned mode vector when the high-level policy determines that a high-dynamic execution regime is appropriate.

\subsection{Two-Stage Training and Curriculum Learning}
\label{subsec: Two-Stage Training and Curriculum Learning}

We adopt a decoupled two-stage training paradigm and employ the Proximal Policy Optimization (PPO) algorithm ~\cite{schulman2017proximal} during training. The training stage is based on the NVIDIA Isaac Lab\cite{mittal2025isaac} simulator.

\begin{figure}[!t]
    \centering
    \includegraphics[width=0.65\columnwidth]
    {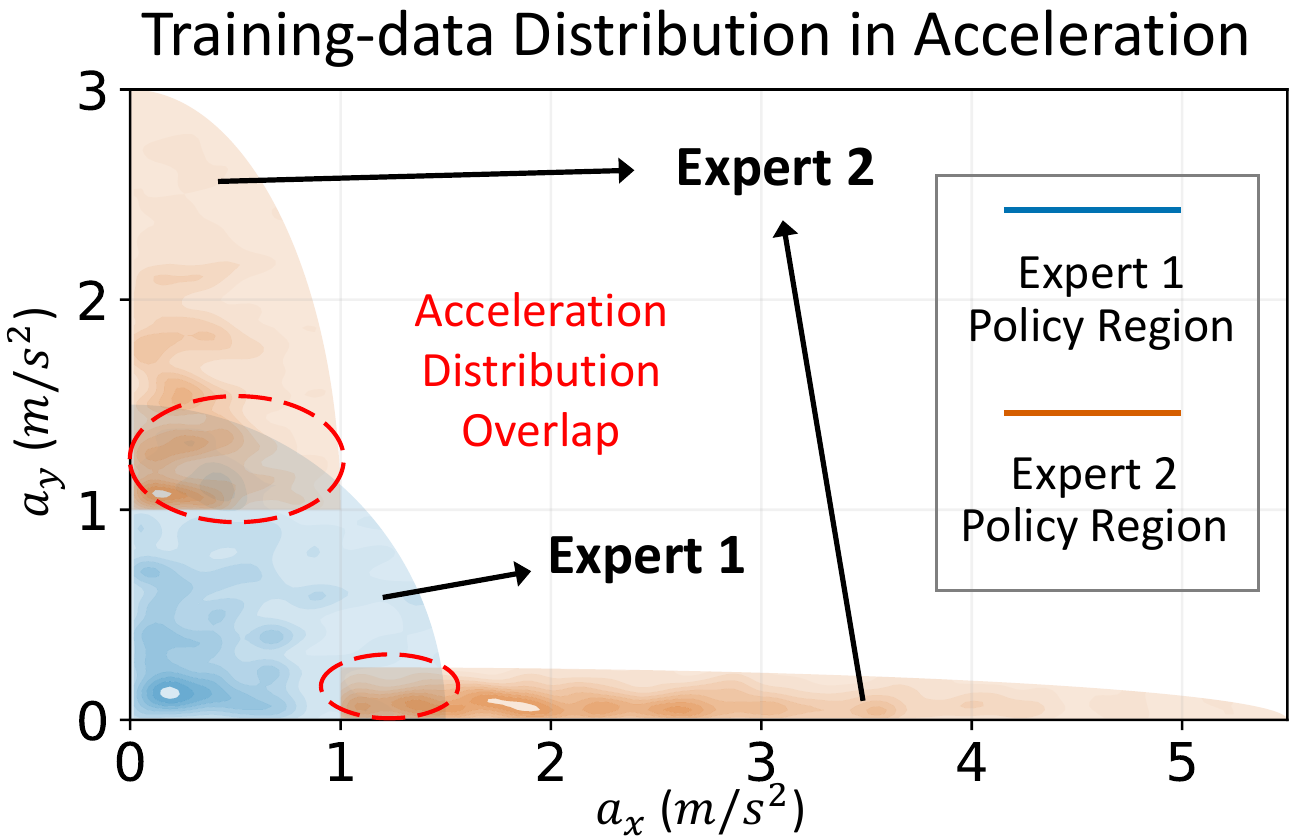}
    \caption{The training acceleration distribution for two experts. \textbf{Expert 1:} Low-Speed Omnidirectional Network 
    ($\|a\| \leq 1.5\,\text{m}/\text{s}^2$)
, \textbf{Expert 2:} High-Dynamic Agile Network
($\|a_x\| \in [1.0, 5.5] \, \text{m/s}^2$,
$\|a_y\| \in [1.0, 3.0] \, \text{m/s}^2$).
}
    \label{Fig: data_distribution}
\vspace{-2mm}
\end{figure}

\textbf{Pre-training the Low-Level Dual-Mode Controller.} 
To establish robust velocity tracking capabilities in the first stage, the two expert networks $\pi_{\text{low}}$ and $\pi_{\text{high}}$ are trained independently to master their respective operational domains.
The primary objective for training Low-Speed Omnidirectional Expert ($\pi_{\text{low}}$) is precise omnidirectional command tracking. We randomly sample velocity commands within a low-speed navigation range ($v_x, v_y$). Through training, the policy learns to adaptively adjust its gait, naturally emerging rolling, stepping, and hybrid gait modes (Fig. \ref{Fig: data_distribution}).
High-Dynamic Agile Expert ($\pi_{\text{high}}$) is trained for extreme evasive maneuvers, focusing on high-acceleration startups in both rolling and stepping gaits (Fig. \ref{Fig: data_distribution}). To ensure capability to brake safely after a high-speed rolling sprint, we explicitly introduce the historical command sequence $\mathbf{v}_{\text{cmd}}^{t-4:t}$. If the sequence indicates a sudden deceleration intent, the network learns to proactively adjust the base pitch and modulate wheel damping to achieve stable and rapid braking.

\textbf{High-Level Avoidance Policy with Curriculum Learning.} 
We progressively increase the flight speed of the dynamic obstacles based on the evasion success rate.
We formulate the reward function based on two primary principles: maximizing task success and ensuring physical feasibility. The total reward $r_{t}$ is a weighted sum of the following components:

\textbf{Task Success and Safety Margin.} We prioritize the ultimate evasion outcome over dense spatial tracking constraints (Sec. \ref{Sec: prelim}). The policy is rewarded for successful evasion with a guaranteed safety margin, while collisions are strictly penalized.
Beyond the safety-margin-based task objective, we intentionally include a proportion of static or near-static obstacles, as well as non-centrally moving obstacles with no effective collision threat, to regularize the policy against false-positive evasion. For these non-threatening scenarios, an explicit penalty is imposed whenever the system produces an avoidance response.
Let $d_{\text{min}}$ be the minimum signed surface clearance between the robot collision geometry and the obstacle over the episode, and $d_{\text{safe}}$ be the defined safety threshold:
\begin{equation}
\footnotesize
r_{\text{task}} =
\begin{cases}
R_{\text{succ}}, & \text{if } d_{\min} > d_{\text{safe}} \text{ and } \xi = 1 \text{ (Success)} \\[4pt]
R_{\text{fail}}, & \text{if } 0 < d_{\min} \le d_{\text{safe}} \text{ (Collision)} \\[4pt]
-\lambda_{\text{fp}} \, \|\text{v}_{\text{cmd}}\|^2, & \text{if } \xi = 0 \text{ and } \alpha_t = 1 \text{ as defined in } \eqref{eq:threat_activation} \\[4pt]
0, & \text{otherwise}
\end{cases},
\label{eq:task_reward_fp}
\end{equation}
where $R_{\text{succ}} > 0$ and $R_{\text{fail}} < 0$, $\lambda_{\text{fp}} > 0$ is the false-positive penalty weight, and $\xi \in \{0,1\}$ is a threat indicator, 
$\xi =
\mathbb{I}\!\left(
\kappa > 0
\;\land\;
d_{\min}^{\mathrm{pred}} < d_{\mathrm{safe}}
\right)$,
where $d_{\min}^{\mathrm{pred}}$ denotes the predicted minimum clearance between the robot and the obstacle within the prediction horizon,
$\xi=1$ denotes a predicted threat requiring an evasive response, while $\xi=0$ denotes a non-threatening obstacle.
We set $R_{\text{succ}}=5$, $R_{\text{fail}}=-10$, $\lambda_{\text{fp}}=0.10$, and $d_{\text{safe}}=0.15\,\mathrm{m}$.

\textbf{Dynamic Feasibility and Efficiency Regularization.} To ensure the high-level commands $\mathbf{v}_{\text{cmd}}$ respect the non-holonomic constraints of the wheeled-legged morphology and maintain smooth, energy-efficient operations, we introduce a composite regularization term. First, if the frozen low-level controller fails to track the command, the command is deemed physically invalid, penalized by a tracking error term $r_{\text{track}} = -k_{\text{track}} \|\mathbf{v}_{\text{cmd}} - \mathbf{v}_{\text{real}}\|^2$. Second, to prevent erratic mode switching and ensure trajectory smoothness, we penalize velocity jumps via $r_{\text{smooth}} = -k_{\text{smooth}} \|\mathbf{v}_{\text{cmd}}^t - \mathbf{v}_{\text{cmd}}^{t-1}\|^2$. Finally, we encourage energy efficiency,
$r_{\text{energy}} = - k_{\text{power}} \sum_{i=1}^{n_j} |\tau_i \cdot \dot{q}_i|
$,
where $\tau_i$ and $\dot{q}_i$ are the joint torques and velocities.
The regularization weights are set as $[k_{\text{track}},k_{\text{smooth}},k_{\text{power}}]=[0.50,0.05,10^{-4}]$.

\subsection{Domain Randomization for Sim-to-Real Transfer}

\begin{table}[t]
\caption{Domain Randomization for Sim-to-Real Transfer}
\label{tab:domain_randomization}
\centering
\footnotesize
\setlength{\tabcolsep}{2pt}
\renewcommand{\arraystretch}{1.12}
\begin{tabularx}{\columnwidth}{@{}
    >{\raggedright\arraybackslash}p{0.15\columnwidth}
    >{\raggedright\arraybackslash}p{0.49\columnwidth}
    >{\raggedright\arraybackslash}X
@{}}
\toprule
\textbf{Group} & \textbf{Parameter} & \textbf{Range} \\
\midrule

\multirow{2}{*}{Obstacle}
& Initial distance $d_{\mathrm{obs}}$
& $[3.0,\,5.0]~\mathrm{m}$ \\
& Still / non-central ratio
& $0.05$ \\

\multirow{3}{*}{Base init.}
& Height $z_0$
& $[0,\,0.25]~\mathrm{m}$ \\
& Roll / pitch $(\theta_0,\phi_0)$
& $[-0.3,\,0.3]~\mathrm{rad}$ \\
& Angular velocity $\omega_0$
& $[-0.5,\,0.5]~\mathrm{rad\,s^{-1}}$ \\

\multirow{3}{*}{Actuation}
& Stiffness scale $k_{\mathrm{stiff}}$
& $[0.7,\,1.3]$ \\
& Damping scale $k_{\mathrm{damp}}$
& $[0.7,\,1.3]$ \\
& External force / torque $(f_{\mathrm{ext}}, \tau_{\mathrm{ext}})$
& $[-10,\,10]~\mathrm{N},~\mathrm{N\!\cdot\!m}$ \\

\multirow{5}{*}{Body}
& COM offset $p_{\mathrm{com}}$
& $[-0.03,\,0.03]~\mathrm{m}$ \\
& Inertia scale $I_{\mathrm{rb}}$
& $[0.5,\,1.5]$ \\
& Base-mass offset $\Delta m_b$
& $[-1,\,3]~\mathrm{kg}$ \\
& Limb-mass scale $m_l$
& $[0.7,\,1.3]$ \\
& Friction coefficient $\mu$
& $[0.3,\,0.8]$ \\
\bottomrule
\end{tabularx}
\vspace{-1mm}
\end{table}

To facilitate sim-to-real transfer, we apply structured domain randomization during training to improve robustness against modeling discrepancies in robot dynamics, body parameters, actuation, and obstacle conditions, as summarized in Table~\ref{tab:domain_randomization}.
The randomized training environment also includes static or near-static obstacles and non-central moving obstacles with varying approach directions and speeds. These perturbations expose the policy to both benign and threatening obstacle states and regularize unnecessary evasive responses.

%% file: sections/5_Experiment.tex
\section{Simulation Experiments} \label{Sec: simulation}

This section aims to evaluate the agility and efficiency of the proposed \textbf{AWARE} framework in dynamic evasion tasks. 
In particular, we investigate the behavioral modes elicited by the robot’s hybrid morphology, analyze how the framework responds to obstacles with different approach patterns, and assess its extreme evasion capability under highly dynamic threats.

\begin{figure*}[t]
    \centering
    \includegraphics[width=0.90\textwidth]{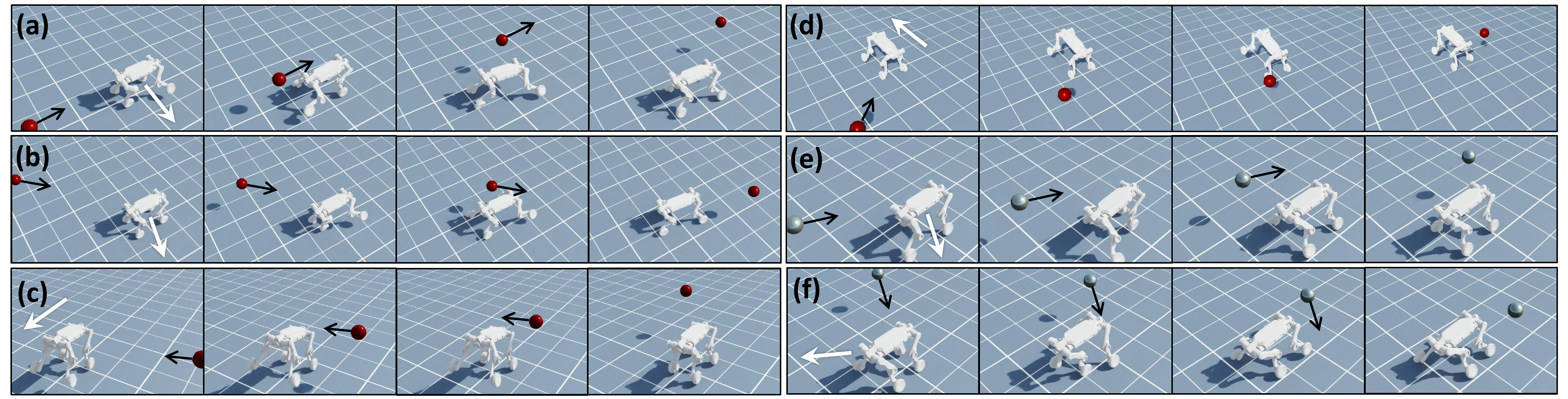}
    \vspace{-1mm}
    \caption{\textbf{Visualization of the dual-mode high-dynamic obstacle avoidance system.} \textbf{Reflexive Evasion Mode (a--d):} Triggered by high-speed obstacles (red ball) from varying directions, the system exhibits extreme maneuvers, specifically lateral jumping within the stepping mode (a, b) and forward leaping within the rolling mode (c, d). \textbf{Navigation Avoidance Mode (e--f):} Under lower threat levels (silver ball), the robot executes smooth and continuous avoidance utilizing stepping and hybrid gaits.}
    \label{Fig: simulation_demo}
    \vspace{-5mm}
\end{figure*}

\subsection{Simulation Settings}

\textbf{Task Setup.}
The simulation experiments are conducted within the IsaacLab framework~\cite{mittal2025isaac}. To evaluate the system's dynamic avoidance capabilities, we design an omnidirectional obstacle-dodging scenario. 
Specifically, a spherical obstacle is spawned at an initial distance $d_0 \sim \mathcal{U}(3,5)\,\mathrm{m}$ from the robot's center of mass and moves horizontally toward the robot. Its approach direction is selected from a discrete set of 13 predefined angles spanning $0^\circ$ to $180^\circ$, while its velocity is determined by the prescribed available reaction time $t_r \in [0.0,3.5]\,\mathrm{s}$.
For balanced statistics over reaction time and obstacle direction, we divide $t_r$ into $0$--$1$, $1$--$2$, and $2$--$3\,\mathrm{s}$ intervals, with five reaction-time points per interval. Each point is evaluated over 13 predefined approach directions and 20 trials per direction, distributed across five independently trained seeds with four repetitions per seed.
The training hyperparameters are summarized in Table~\ref{tab:training_hyperparameters}.

\begin{table}[t]
\centering
\caption{TRAINING AND NETWORK HYPERPARAMETERS}
\label{tab:training_hyperparameters}
\scriptsize
\renewcommand{\arraystretch}{1.05}
\setlength{\tabcolsep}{1pt}
    \resizebox{\columnwidth}{!}{%
\begin{tabular}{@{}lclc@{}}
\toprule
\textbf{Optimization} & \textbf{Value}
& \textbf{Training Setup} & \textbf{Value} \\
\midrule

Optimizer
& Adam
& $(N_{\mathrm{env}},H_{\mathrm{roll}})$
& $4096, 24$ \\

$(\eta,\gamma,\lambda_{\mathrm{GAE}})$
& $10^{-3}, 0.99, 0.95$
& $(N_{\mathrm{iter}},N_{\mathrm{mb}},N_{\mathrm{ep}})$
& $2000, 4, 5$ \\

$(\epsilon_{\mathrm{clip}},c_v,c_{\mathrm{ent}})$
& $0.20, 1.0, 0$
& Actor MLP
& $[256,256,128]$ \\

$g_{\max}$, LR schedule
& $1.0$, constant
& Activation, $\tau_g$
& ELU, $1.0$ \\

Obs./adv. normalization
& enabled/enabled
& Training seeds
& $5$ \\

\bottomrule
\end{tabular}
}
\vspace{-1mm}
\end{table}

\textbf{Metrics.}
We define the following key metrics:
Avoidance Success Rate (\textbf{ASR}), which is the ratio of successful trials to the total number of experiments, $\text{ASR} = {N_{\text{success}}}/{N_{\text{total}}}$.
Avoidance Maneuver Distance (\textbf{AMD}) is used to measure the spatial efficiency of the evasion maneuver, defined as the Euclidean distance between the base's initial and final positions, $\text{AMD} = | \mathbf{p}_{\text{base}}^{\text{final}} - \mathbf{p}_{\text{base}}^{\text{init}} |$.

\textbf{Baselines.}
We selected two representative baseline algorithms for comparative experiments: \textbf{REBot}~\cite{xu2025rebot} (locked wheels) proposes a reflexive evasion framework that handles fast-moving threats across various approach angles.
 The Relaxed Barrier Function MPC (\textbf{RB-MPC})~\cite{gaertner2021collision} enables legged robots to anticipate future collisions and automatically execute complex, dynamic avoidance maneuvers. For RB-MPC, we use a prediction horizon of $1\,\mathrm{s}$ and a replanning frequency of $20\,\mathrm{Hz}$.
All methods are evaluated on the same M20 model using identical initial states, obstacle trajectories, state measurements, collision criteria, and actuation limits.

\subsection{Main Experimental Results}

As illustrated in Fig. \ref{Fig: simulation_demo}, without imposing dense heuristic action constraints, the system is driven purely by the high-level obstacle avoidance objective. Depending on the perceived threat level, the policy autonomously selects the optimal reaction velocity.
Specifically, for navigation avoidance, the robot performs continuous transitions among stepping, rolling, and hybrid gaits to maintain smooth and energy-efficient trajectories. Under highly dynamic reflexive evasion, the system exhibits extreme maneuvers such as lateral jumping (Fig.  \ref{Fig: simulation_demo}(a)(b)) and aggressive forward leaping (Fig.  \ref{Fig: simulation_demo}(c)(d)). Notably, 
during high-speed rolling maneuvers, it learns to autonomously lock the wheel motors upon generating sharp deceleration commands, instantly transitioning to a high-friction state to promote rapid and dynamically stable braking.

\begin{figure}[!t]
    \centering
    \includegraphics[width=0.98\columnwidth]{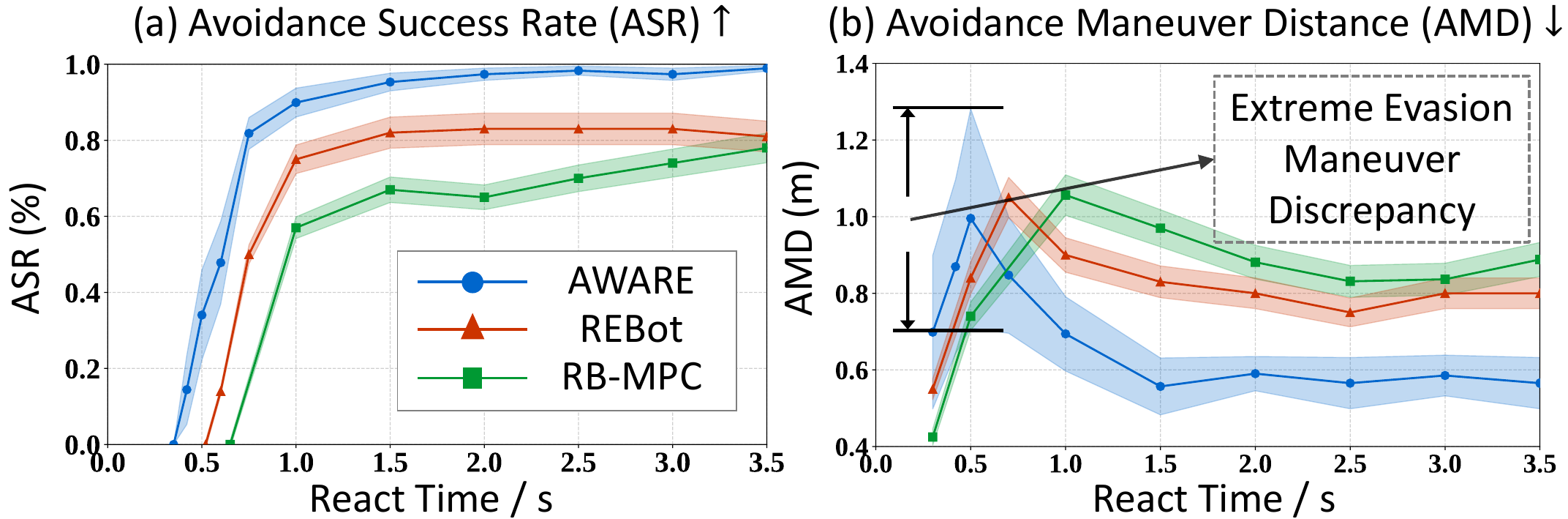}
    \vspace{-1mm}
    \caption{
    Performance comparison of the proposed method and baseline methods for high-dynamic obstacle avoidance: AWARE, REBot, and RB-MPC.
    }
    \label{fig:baseline_high_dynamic}
    \vspace{-1mm}
\end{figure}

\begin{table}[htbp]
    \centering
    \renewcommand{\arraystretch}{1.1}
    \caption{AWARE results across reaction-time intervals}
    {\footnotesize  
    \resizebox{\columnwidth}{!}{
    \begin{threeparttable}
    \setlength{\tabcolsep}{6pt}
    \begin{tabular}{c c c c}
        \toprule
        \textbf{Reaction Time (s)} & \textbf{ASR Mean $\pm$ SD} & \textbf{95\% CI} & \textbf{AMD Mean $\pm$ SD} \\
        \midrule
        0--1 & $0.294 \pm 0.315$ & $[0.270,0.319]$ & $0.751 \pm 0.246$ \\
        1--2 & $0.931 \pm 0.058$ & $[0.916,0.943]$ & $0.626 \pm 0.129$ \\
        2--3 & $0.973 \pm 0.027$ & $[0.963,0.981]$ & $0.580 \pm 0.041$ \\
        \bottomrule
    \end{tabular}
    \end{threeparttable}
    }}
    \vspace{1mm}
\end{table}

The results in Fig. ~\ref{fig:baseline_high_dynamic} indicate that AWARE maintains an overall performance advantage across different reaction-time conditions, particularly under constrained reaction windows. RB-MPC exhibits limitations in high-dynamic scenarios: its performance degrades due to fast-changing environments and model mismatches, which hinder timely responses to rapidly approaching disturbances. Similarly, REBot fails to achieve stable obstacle avoidance on the wheeled-legged platform, as it does not adequately account for the unique dynamics of the wheeled-legged system, resulting in substantially lower success rates. We further evaluate the sensitivity of $v_{\mathrm{th}}$ and $d_{\mathrm{th}}$ by varying them within $\pm 20\%$ of their nominal values. The average ASR remains within $72\%$--$76\%$, suggesting stable performance under moderate threshold variations in the tested scenarios, rather than general robustness to erroneous threat activations.

\subsection{Ablation Studies}

\begin{figure}[htbp]
    \centering
\includegraphics[width=0.95\columnwidth]{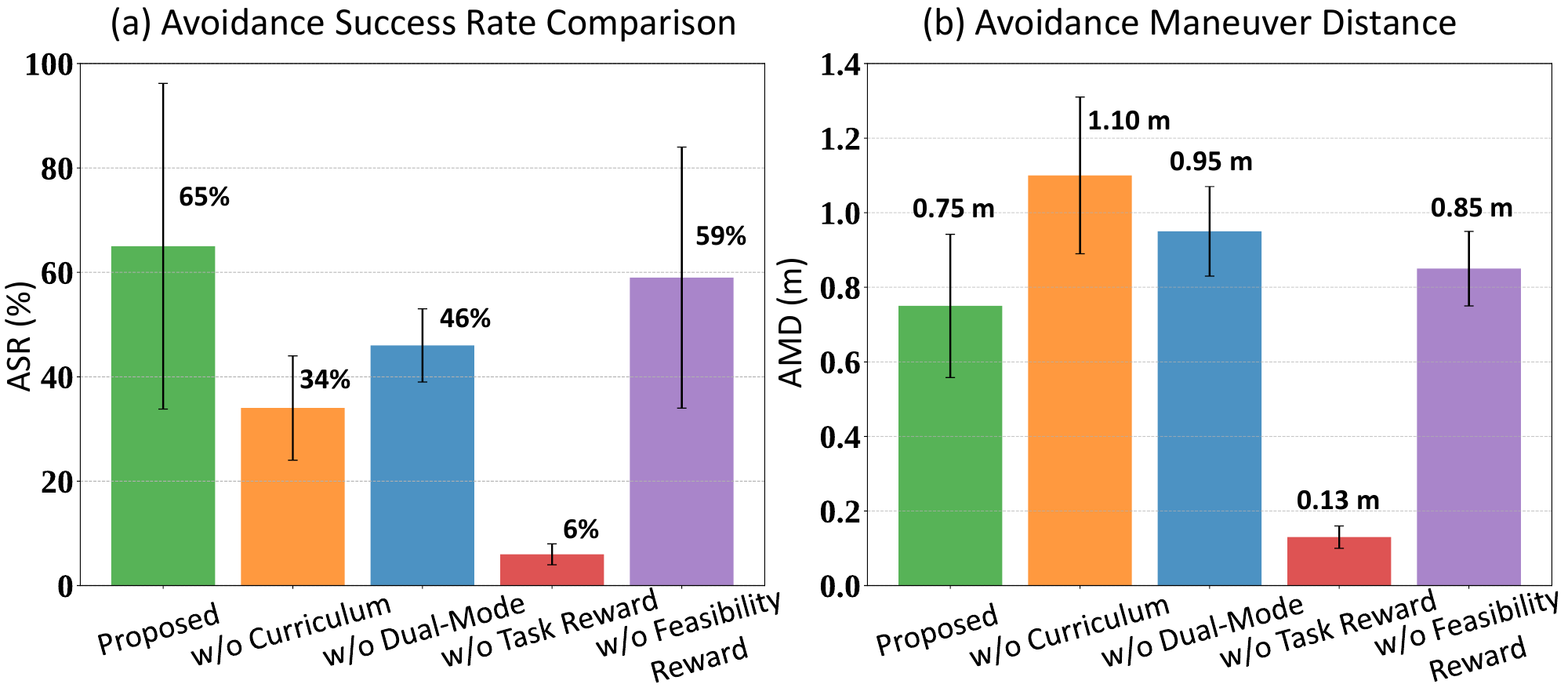}
    \vspace{-1mm}
    \caption{Ablation studies on the AWARE framework. (a) Avoidance Success Rate (ASR). (b) Avoidance Maneuver Distance (AMD) to the obstacle.}
    \label{fig:ablation}
\vspace{-3mm}
\end{figure}

We ablate curriculum learning, dual-mode specialization, task reward, and feasibility regularization. System performance is measured primarily via the Avoidance Success Rate (ASR) (Fig. ~\ref{fig:ablation}(a)) and the Avoidance Maneuver Distance (AMD) (Fig. ~\ref{fig:ablation}(b)) to the obstacle. The ablation results confirm the complementary contribution of each component. Removing dual-mode specialization causes mode confusion, reducing agility and leading to balance failures under extreme threats. The \emph{w/o Curriculum} variant converges more slowly and shows degraded overall performance. Removing the task reward eliminates the explicit evasion incentive and causes a sharp drop in ASR, whereas removing feasibility regularization permits physically infeasible velocity commands that may occasionally avoid obstacles but frequently induce low-level tracking failures.

\subsection{Cruise--Evade--Resume Transition}

\begin{figure}[H]
    \centering
    \includegraphics[width=\columnwidth]{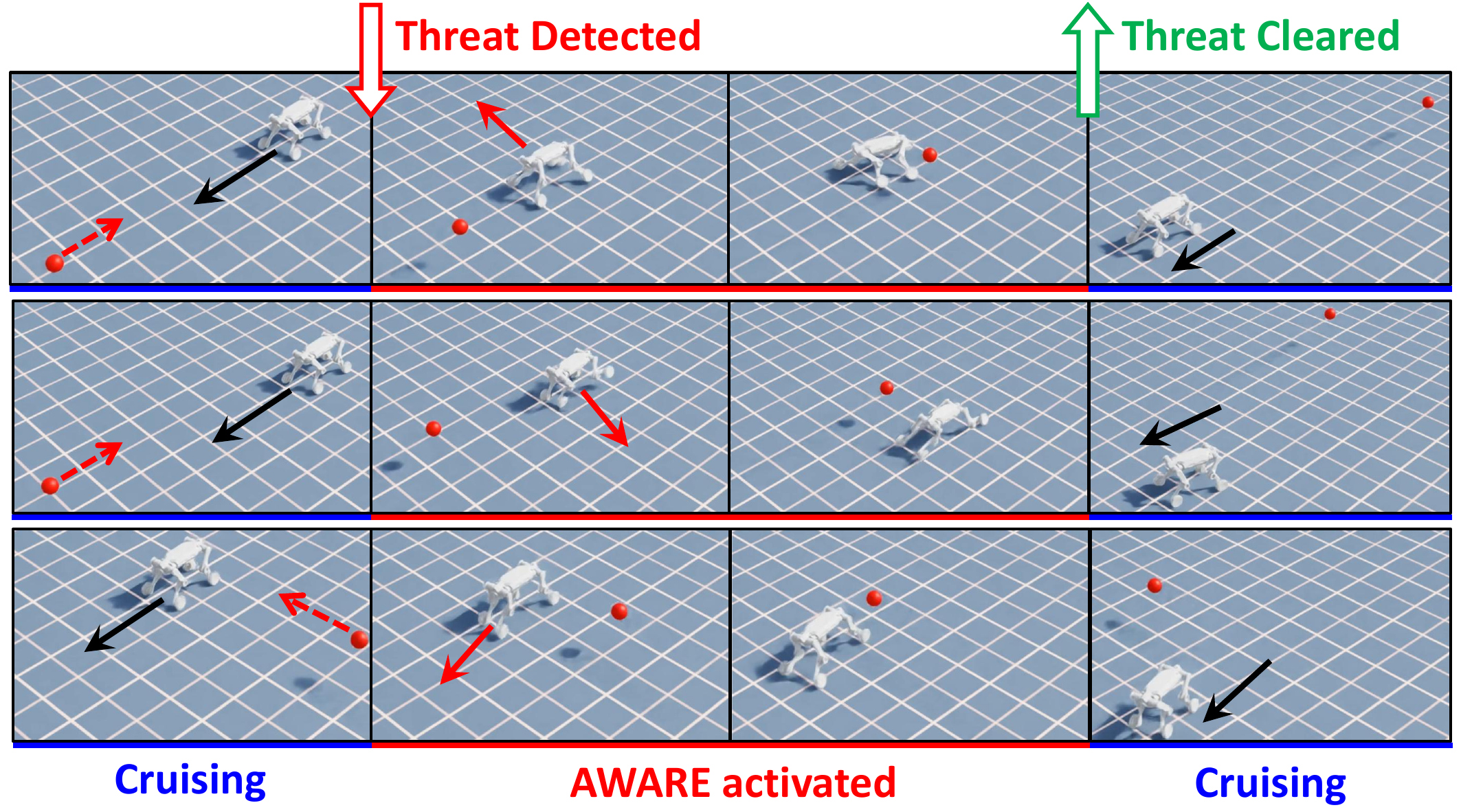}
    \vspace{-4mm}
    \caption{Cruise--evade--resume experiment for online transition validation. The robot first executes navigation-avoidance, switches to reflexive-evasion when a dynamic threat enters the activation region, and then resumes navigation-avoidance after the obstacle has passed.}
    \label{fig:cruise_evade}
    \vspace{-1mm}
\end{figure}

\begin{figure*}[t]
    \centering
    \includegraphics[width=0.80\textwidth]{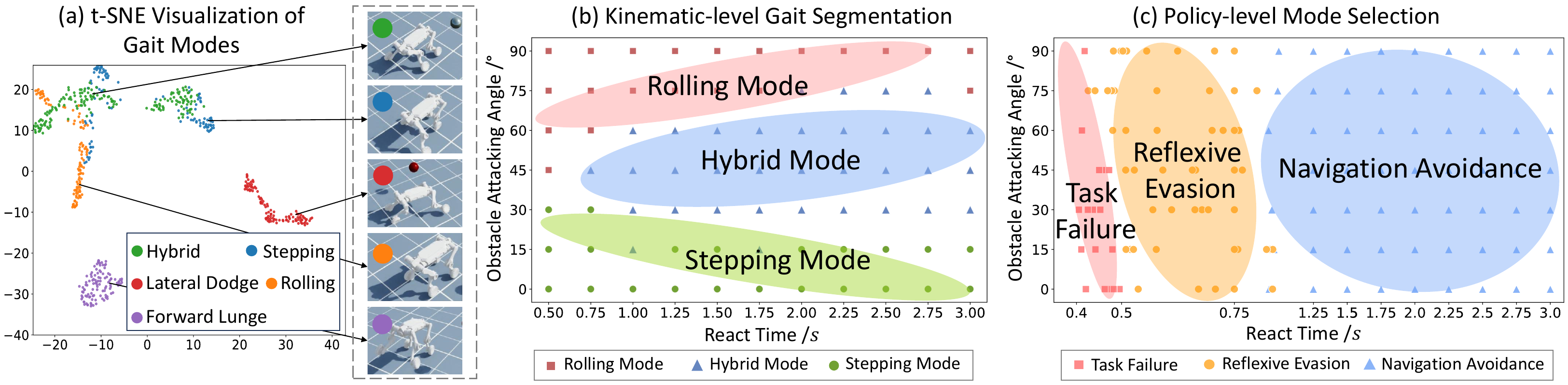}
    \caption{\textbf{Behavioral analysis of AWARE.} (a) t-SNE visualization of emergent locomotion modes. (b) Gait distribution under different reaction times and obstacle angles. (c) High-level mode selection between navigation avoidance and reflexive evasion.}
    \label{Fig: evaluation_categorization}
\vspace{-1mm}
\end{figure*}

To evaluate AWARE during ongoing navigation-avoidance, we conduct cruise--evade--resume experiments at a cruising speed of $0.2\,\mathrm{m/s}$, with obstacles approaching at a reaction time of $0.75\,\mathrm{s}$, as shown in Fig.~\ref{fig:cruise_evade}. Across 20 frontal and 20 lateral trials, AWARE achieves an average success rate of $72.5\%$, demonstrating online takeover from navigation-avoidance to reflexive-evasion and return after evasion.

\subsection{Extreme Dynamic Evasion and Gait Modes}

We evaluate extreme start-up responses at $t_r=0.42\,\mathrm{s}$ and $0.57\,\mathrm{s}$ along the $x$- and $y$-directions, respectively.
Start-up performance is quantified using Transient Escape Acceleration (TEA) and Torque-Normalized Acceleration (TNA):
\begin{equation}
\small
\mathrm{TEA} = \frac{1}{\Delta t_{\mathrm{start}}} \int_{t_{\mathrm{detect}}}^{t_{\mathrm{detect}} + \Delta t_{\mathrm{start}}} a(t)\,dt,
\label{eq:tea}
\end{equation}
\begin{equation}
\small
\mathrm{TNA}
=
\frac{\bar{a}_{\mathrm{transient}}}
{\left(\bar{\tau}_{\mathrm{transient}} / \bar{\tau}_{\mathrm{idle}}\right)},
\label{eq:nte}
\end{equation}

Table~\ref{tab:startup_dynamics} shows that rolling-based start-up achieves higher TEA and TNA, with mechanical power concentrated in wheel and hip--$Y$ actuation, whereas stepping-based start-up relies more on the hip--$X$ and knee joints. These results highlight their distinct actuation patterns and the stronger transient response of rolling-based evasion.

\begin{figure*}[t]
    \centering
    \includegraphics[width=0.75\textwidth]{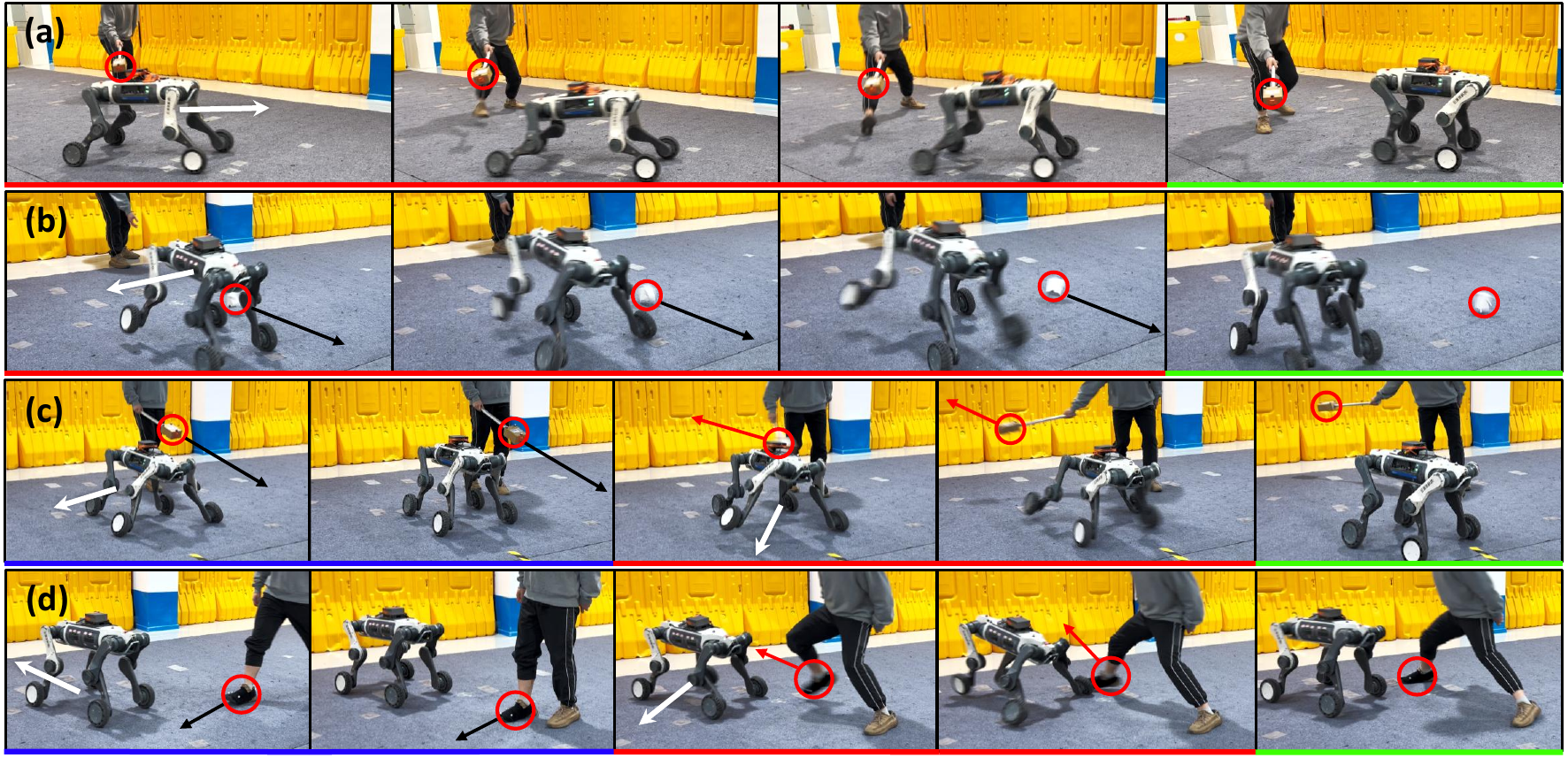}
    \caption{\textbf{Real-robot experiments for high-dynamic obstacle avoidance.}
    (a) Reflexive evasion of a y-direction obstacle using a rolling gait.
    (b) Reflexive evasion of an x-direction obstacle using a stepping gait.
    (c--d) Continuous mixed-mode evasion for obstacles with varying speeds, initiating with navigation avoidance and transitioning to reflexive evasion.
    Red represents \textcolor{red}{the reflexive evasion phase}, blue represents the navigation avoidance phase, and green represents \textcolor{green}{the standing recovery phase}.}
    \label{fig:real_robot}
    \vspace{-4mm}
\end{figure*}

\begin{table}[htbp]
    \centering
    \caption{Start-up dynamics of rolling and stepping modes.}
    \label{tab:startup_dynamics}
    \setlength{\tabcolsep}{5pt}
    \renewcommand{\arraystretch}{0.92}
    \begin{tabular}{llcc}
        \toprule
        \textbf{Category} & \textbf{Metric}
        & \textbf{Rolling} & \textbf{Stepping} \\
        \midrule

        \multirow{2}{*}{Start-up dynamics}
        & TEA ($\mathrm{m/s^2}$) & 4.13 & 2.78 \\
        & TNA ($-$)              & 3.678 & 1.251 \\

        \midrule

        \multirow{4}{*}{Actuator power (W)}
        & Wheel   & 485.92 & 235.14 \\
        & Hip--$X$ & 6.83  & 84.84 \\
        & Hip--$Y$ & 50.93 & 37.24 \\
        & Knee    & 31.13  & 44.26 \\

        \bottomrule
    \end{tabular}
    \vspace{-4mm}
\end{table}

\textbf{We visualize the learned kinematic feature space} using t-SNE to verify that AWARE induces multiple behaviorally distinct locomotion and evasive modes. As shown in Fig.~\ref{Fig: evaluation_categorization}(a), the t-SNE embedding separates the five observed behaviors: Hybrid bridges the compact Rolling and Stepping clusters, while Forward Lunge and Lateral Dodge form distinct high-speed regions. \textbf{We systematically evaluate the system's behavioral responses by introducing a two-level categorization scheme}. First, at the kinematic level, we cluster the naturally emerged locomotion behaviors into three distinct gait categories (Fig. ~\ref{Fig: evaluation_categorization}(b)): \textit{Stepping}, \textit{Rolling}, and \textit{Hybrid}. Second, at the policy execution level, we classify the system's operational states into two primary modes based on the high-level policy's selection of the low-level experts: \textit{Navigation Avoidance} and \textit{Reflexive Evasion} (Fig. \ref{Fig: evaluation_categorization}(c)). We examine how the robot adaptively transitions between these modes and gaits under varying available reaction times and diverse obstacle approach angles.

\section{Real-Robot Demonstration}
We validate the proposed framework in real-world experiments on the M20 wheeled-legged robot platform. A motion-capture system provides real-time position and velocity measurements for both the robot and the dynamic obstacle for state estimation and evaluation (Fig. \ref{fig:real_robot}).
To emulate dynamic obstacles, we consider three representative interaction settings: box throwing, stick poking, and foot kicks. These disturbances are introduced from multiple directions, including the front, left, right, and diagonal directions, in order to evaluate the learned policy across representative interaction types and approach directions.

The real-robot experimental results demonstrate that AWARE autonomously adapts its evasive response to the incoming obstacle state and returns the robot to a stable posture after the maneuver.
Across 75 real-robot reflexive evasion trials, the AWARE system achieves an average avoidance success rate (ASR) of $71\%$ and an avoidance maneuver distance (AMD) of $1.2\,\mathrm{m}$.
Failure analysis indicates that the main failure modes arise from wheel-velocity or joint-torque saturation and conservative posture-safety clipping, latency in the motion-capture and communication pipeline, and residual simulation-to-real discrepancies in wheel-ground friction, actuator response, and robot dynamics.
We further test the framework in continuous mixed-evasion scenarios (Fig. \ref{fig:real_robot}(c)(d)) that integrate reflexive evasion with navigation-aware avoidance. In these experiments, the robot exhibits both navigation-avoidance and reflexive-evasion behaviors according to obstacle speed, demonstrating online mode switching on the physical platform.

%% file: sections/6_Conclusion.tex
\section{Conclusion} \label{Sec: conclusion}

In this paper, we presented AWARE, a hierarchical reinforcement learning framework that enables high-dynamic reflexive obstacle evasion for wheeled-legged robots. Simulation and real-world experiments on the M20 platform demonstrate that the proposed framework can generate diverse evasive behaviors, including rolling-, stepping-, and hybrid-dominated maneuvers, while achieving online transitions under dynamic interactions. This work reveals the agility potential of wheeled-legged robots beyond conventional locomotion and navigation tasks, showing that their complementary rolling and stepping capabilities can be effectively leveraged for rapid reactive behaviors. Future work will investigate extending the current single-threat reflexive module into a comprehensive safety-aware navigation framework with richer perception and multi-objective decision making.

%% file: references.bib
@article{hwangbo2019learning,
  title={Learning agile and dynamic motor skills for legged robots},
  author={Hwangbo, Jemin and Lee, Joonho and Dosovitskiy, Alexey and Bellicoso, Dario and Tsounis, Vassilios and Koltun, Vladlen and Hutter, Marco},
  journal={Science Robotics},
  volume={4},
  number={26},
  pages={eaau5872},
  year={2019},
  publisher={American Association for the Advancement of Science}
}

@article{miki2022learning,
  title={Learning robust perceptive locomotion for quadrupedal robots in the wild},
  author={Miki, Takahiro and Lee, Joonho and Hwangbo, Jemin and Wellhausen, Lorenz and Koltun, Vladlen and Hutter, Marco},
  journal={Science Robotics},
  volume={7},
  number={62},
  pages={eabk2822},
  year={2022},
  publisher={American Association for the Advancement of Science}
}

@article{lee2024learning,
  title={Learning robust autonomous navigation and locomotion for wheeled-legged robots},
  author={Lee, Joonho and Bjelonic, Marko and Reske, Alexander and Wellhausen, Lorenz and Miki, Takahiro and Hutter, Marco},
  journal={Science Robotics},
  volume={9},
  number={89},
  pages={eadi9641},
  year={2024},
  publisher={American Association for the Advancement of Science}
}

@article{zhang2025dynamic,
  title={Dynamic obstacle avoidance for car-like mobile robots based on neurodynamic optimization with control barrier functions},
  author={Zhang, Zheng and Yang, Guang-Hong},
  journal={Neurocomputing},
  volume={654},
  pages={131252},
  year={2025},
  publisher={Elsevier},
  doi={10.1016/j.neucom.2025.131252}
}

@article{zhou2023max,
  title={Max: A wheeled-legged quadruped robot for multimodal agile locomotion},
  author={Zhou, Qinqin and Yang, Sicheng and Jiang, Xinyang and Zhang, Dongsheng and Chi, Wanchao and Chen, Ke and Zhang, Shenghao and Li, Jie and Zhang, Jingfan and Wang, Rui and others},
  journal={IEEE Transactions on Automation Science and Engineering},
  volume={21},
  number={4},
  pages={7562--7582},
  year={2024},
  publisher={IEEE}
}

@article{chen2025gmt,
  title={{GMT}: General motion tracking for humanoid whole-body control},
  author={Chen, Zixuan and Ji, Mazeyu and Cheng, Xuxin and Peng, Xuanbin and Peng, Xue Bin and Wang, Xiaolong},
  journal={arXiv preprint arXiv:2506.14770},
  year={2025}
}

@inproceedings{rudin2022advanced,
  title={Advanced skills by learning locomotion and local navigation end-to-end},
  author={Rudin, Nikita and Hoeller, David and Bjelonic, Marko and Hutter, Marco},
  booktitle={2022 IEEE/RSJ International Conference on Intelligent Robots and Systems (IROS)},
  pages={2497--2503},
  year={2022},
  organization={IEEE}
}

@article{bjelonic2020rolling,
  title={Rolling in the deep--hybrid locomotion for wheeled-legged robots using online trajectory optimization},
  author={Bjelonic, Marko and Sankar, Prajish K and Bellicoso, C Dario and Vallery, Heike and Hutter, Marco},
  journal={IEEE Robotics and Automation Letters},
  volume={5},
  number={2},
  pages={3626--3633},
  year={2020},
  publisher={IEEE}
}

@inproceedings{huang2025moe,
  title={{MoE-Loco}: Mixture of experts for multitask locomotion},
  author={Huang, Runhan and Zhu, Shaoting and Du, Yilun and Zhao, Hang},
  booktitle={2025 IEEE/RSJ International Conference on Intelligent Robots and Systems (IROS)},
  pages={14218--14225},
  year={2025},
  organization={IEEE}
}

@article{xu2025rebot,
  title={{REBot}: Reflexive Evasion Robot for Instantaneous Dynamic Obstacle Avoidance},
  author={Xu, Zihao and Hao, Ce and Wang, Chunzheng and Sima, Kuankuan and Shi, Fan and Dong, Jin Song},
  journal={IEEE Robotics and Automation Letters},
  year={2026},
  volume={11},
  number={2},
  pages={1746--1753},
  doi={10.1109/LRA.2025.3643278}
}

@article{xu2025intent,
  title={Intent prediction-driven model predictive control for {UAV} planning and navigation in dynamic environments},
  author={Xu, Zhefan and Jin, Hanyu and Han, Xinming and Shen, Haoyu and Shimada, Kenji},
  journal={IEEE Robotics and Automation Letters},
  volume={10},
  number={5},
  pages={4946--4953},
  year={2025},
  publisher={IEEE},
  doi={10.1109/LRA.2025.3555937}
}

@article{mittal2025isaac,
  title={Isaac {Lab}: A {GPU}-accelerated simulation framework for multi-modal robot learning},
  author={Mittal, Mayank and Roth, Pascal and Tigue, James and Richard, Antoine and Zhang, Octi and Du, Peter and Serrano-Mu{\~n}oz, Antonio and Yao, Xinjie and Zurbr{\"u}gg, Ren{\'e} and Rudin, Nikita and others},
  journal={arXiv preprint arXiv:2511.04831},
  year={2025}
}

@article{zhou2020ego,
  title={{EGO-Planner}: An {ESDF}-free gradient-based local planner for quadrotors},
  author={Zhou, Xin and Wang, Zhepei and Ye, Hongkai and Xu, Chao and Gao, Fei},
  journal={IEEE Robotics and Automation Letters},
  volume={6},
  number={2},
  pages={478--485},
  year={2021},
  publisher={IEEE},
  doi={10.1109/LRA.2020.3047728}
}

@article{burgess2024reactive,
  title={Reactive base control for on-the-move mobile manipulation in dynamic environments},
  author={Burgess-Limerick, Ben and Haviland, Jesse and Lehnert, Chris and Corke, Peter},
  journal={IEEE Robotics and Automation Letters},
  volume={9},
  number={3},
  pages={2048--2055},
  year={2024},
  publisher={IEEE}
}

@article{hoeller2024anymal,
  title={{ANYmal} parkour: Learning agile navigation for quadrupedal robots},
  author={Hoeller, David and Rudin, Nikita and Sako, Dhionis and Hutter, Marco},
  journal={Science Robotics},
  volume={9},
  number={88},
  pages={eadi7566},
  year={2024},
  publisher={American Association for the Advancement of Science}
}

@inproceedings{hosseini2023dynamic,
  title={Dynamic hybrid locomotion and jumping for wheeled-legged quadrupeds},
  author={Hosseini, Mojtaba and Rodriguez, Diego and Behnke, Sven},
  booktitle={2023 IEEE/RSJ International Conference on Intelligent Robots and Systems (IROS)},
  pages={793--799},
  year={2023},
  organization={IEEE}
}

@article{schulman2017proximal,
  title={{Proximal Policy Optimization Algorithms}},
  author={Schulman, John and Wolski, Filip and Dhariwal, Prafulla and Radford, Alec and Klimov, Oleg},
  journal={arXiv preprint arXiv:1707.06347},
  year={2017}
}

@inproceedings{he2024agile,
  title={Agile but safe: Learning collision-free high-speed legged locomotion},
  author={He, Tairan and Zhang, Chong and Xiao, Wenli and He, Guanqi and Liu, Changliu and Shi, Guanya},
  booktitle={Robotics: Science and Systems (RSS)},
  year={2024},
  doi={10.15607/RSS.2024.XX.059}
}

@inproceedings{gaertner2021collision,
  title={Collision-free {MPC} for legged robots in static and dynamic scenes},
  author={Gaertner, Magnus and Bjelonic, Marko and Farshidian, Farbod and Hutter, Marco},
  booktitle={2021 IEEE International Conference on Robotics and Automation (ICRA)},
  pages={8266--8272},
  year={2021},
  organization={IEEE}
}

@inproceedings{chiu2022collision,
  title={A collision-free {MPC} for whole-body dynamic locomotion and manipulation},
  author={Chiu, Jia-Ruei and Sleiman, Jean-Pierre and Mittal, Mayank and Farshidian, Farbod and Hutter, Marco},
  booktitle={2022 IEEE International Conference on Robotics and Automation (ICRA)},
  pages={4686--4693},
  year={2022},
  organization={IEEE}
}

@article{sun2025atros,
  title={{ATRos}: Learning energy-efficient agile locomotion for wheeled-legged robots},
  author={Sun, Jingyuan and Ji, Hongyu and Qu, Zihan and Wang, Chaoran and Zhang, Mingyu},
  journal={arXiv preprint arXiv:2510.09980},
  year={2025}
}

@article{vicente2022fast,
  title={Fast tube model predictive control for driverless cars using linear data-driven models},
  author={Hernandez Vicente, Bernardo A. and Trodden, Paul A. and Anderson, Sean R.},
  journal={IEEE Transactions on Control Systems Technology},
  volume={31},
  number={3},
  pages={1395--1410},
  year={2023},
  publisher={IEEE},
  doi={10.1109/TCST.2022.3224089}
}

@article{shih2023ackerman,
  title={Ackerman unmanned mobile vehicle based on heterogeneous sensor in navigation control application},
  author={Shih, Chi-Huang and Lin, Cheng-Jian and Jhang, Jyun-Yu},
  journal={Sensors},
  volume={23},
  number={9},
  pages={4558},
  year={2023},
  publisher={MDPI}
}

@article{grandia2023perceptive,
  title={Perceptive locomotion through nonlinear model-predictive control},
  author={Grandia, Ruben and Jenelten, Fabian and Yang, Shaohui and Farshidian, Farbod and Hutter, Marco},
  journal={IEEE Transactions on Robotics},
  volume={39},
  number={5},
  pages={3402--3421},
  year={2023},
  publisher={IEEE}
}

@inproceedings{zhang2024learning,
  title={Learning agile locomotion on risky terrains},
  author={Zhang, Chong and Rudin, Nikita and Hoeller, David and Hutter, Marco},
  booktitle={2024 IEEE/RSJ International Conference on Intelligent Robots and Systems (IROS)},
  pages={11864--11871},
  year={2024},
  organization={IEEE}
}

@article{bjelonic2019keep,
  title={Keep rollin’—whole-body motion control and planning for wheeled quadrupedal robots},
  author={Bjelonic, Marko and Bellicoso, C Dario and de Viragh, Yvain and Sako, Dhionis and Tresoldi, F Dante and Jenelten, Fabian and Hutter, Marco},
  journal={IEEE Robotics and Automation Letters},
  volume={4},
  number={2},
  pages={2116--2123},
  year={2019},
  publisher={IEEE}
}

@inproceedings{li2022balancing,
  title={Balancing control and pose optimization for wheel-legged robots navigating high obstacles},
  author={Li, Junheng and Ma, Junchao and Nguyen, Quan},
  booktitle={2022 IEEE/RSJ International Conference on Intelligent Robots and Systems (IROS)},
  pages={8835--8841},
  year={2022},
  organization={IEEE}
}

@inproceedings{schwarz2016hybrid,
  title={Hybrid driving-stepping locomotion with the wheeled-legged robot {Momaro}},
  author={Schwarz, Max and Rodehutskors, Tobias and Schreiber, Michael and Behnke, Sven},
  booktitle={2016 IEEE International Conference on Robotics and Automation (ICRA)},
  pages={5589--5595},
  year={2016},
  organization={IEEE}
}

@article{kashiri2019centauro,
  title={{CENTAURO}: A hybrid locomotion and high power resilient manipulation platform},
  author={Kashiri, Navvab and Baccelliere, Lorenzo and Muratore, Luca and Laurenzi, Arturo and Ren, Zeyu and Hoffman, Enrico Mingo and Kamedula, Malgorzata and Rigano, Giuseppe Francesco and Malzahn, Jorn and Cordasco, Stefano and others},
  journal={IEEE Robotics and Automation Letters},
  volume={4},
  number={2},
  pages={1595--1602},
  year={2019},
  publisher={IEEE}
}

@inproceedings{klemm2019ascento,
  title={Ascento: A two-wheeled jumping robot},
  author={Klemm, Victor and Morra, Alessandro and Salzmann, Ciro and Tschopp, Florian and Bodie, Karen and Gulich, Lionel and K{\"u}ng, Nicola and Mannhart, Dominik and Pfister, Corentin and Vierneisel, Marcus and others},
  booktitle={2019 International conference on robotics and automation (ICRA)},
  pages={7515--7521},
  year={2019},
  organization={IEEE}
}

@inproceedings{wang2021balance,
  title={Balance control of a novel wheel-legged robot: Design and experiments},
  author={Wang, Shuai and Cui, Leilei and Zhang, Jingfan and Lai, Jie and Zhang, Dongsheng and Chen, Ke and Zheng, Yu and Zhang, Zhengyou and Jiang, Zhong-Ping},
  booktitle={2021 IEEE International Conference on Robotics and Automation (ICRA)},
  pages={6782--6788},
  year={2021},
  organization={IEEE}
}
